\documentclass{article}
\usepackage{amssymb}

\usepackage[final]{corl_2025} 
\usepackage{amsmath}
\usepackage{amsfonts}       
\usepackage{microtype}      
\usepackage{bm}
\usepackage{graphicx}
\usepackage{amsfonts}
\usepackage{multicol}
\usepackage{multirow}
\usepackage{booktabs}
\usepackage{graphicx}
\usepackage{array}
\usepackage{bm}
\usepackage{xspace}
\usepackage{float}
\usepackage{wrapfig}
\makeatother
\usepackage{makecell}
\usepackage{nicefrac}
\usepackage{xfrac}
\usepackage{caption}
 \usepackage{enumitem}
\usepackage{adjustbox}
\usepackage{booktabs}
\usepackage{multirow}
\usepackage[ruled]{algorithm2e}
\usepackage{epsfig}
\usepackage{color}
\usepackage{wrapfig,lipsum}
\usepackage{balance}
\usepackage{caption}
\usepackage{multirow}
\usepackage{graphicx}
\usepackage[table,xcdraw]{xcolor}
\usepackage{tocloft}
\hypersetup{
    colorlinks=true,
    linkcolor=blue,
    filecolor=magenta,      
    urlcolor=cyan,
    pdftitle={paperid-208},
    pdfpagemode=FullScreen,
    }
\newlength\savewidth

\definecolor{visual_color}{RGB}{144, 202, 249}
\definecolor{tactile_color}{RGB}{123, 31, 129}
\definecolor{tblblue}{RGB}{31, 119, 180}

\title{VT-Refine: Learning Bimanual Assembly with Visuo-Tactile Feedback via Simulation Fine-Tuning}

\author{
Binghao Huang\textsuperscript{1} \quad Jie Xu\textsuperscript{2} \quad Iretiayo Akinola\textsuperscript{2} \quad Wei Yang\textsuperscript{2} \quad Balakumar Sundaralingam\textsuperscript{2} \\
\textbf{Rowland O'Flaherty\textsuperscript{2}} \quad \textbf{Dieter Fox\textsuperscript{2}} \quad \textbf{Xiaolong Wang\textsuperscript{2,3}} \quad \textbf{Arsalan Mousavian\textsuperscript{2}} \\
\textbf{Yu-Wei Chao\textsuperscript{2$\dagger$}} \quad \textbf{Yunzhu Li\textsuperscript{1$\dagger$}} \\
\textsuperscript{1}Columbia University \quad
\textsuperscript{2}NVIDIA \quad
\textsuperscript{3}University of California, San Diego \\
\textsuperscript{$\dagger$}Equal advising
}

\begin{document}
\maketitle

\vspace{-30pt}
\begin{abstract}
Humans excel at bimanual assembly tasks by adapting to rich tactile feedback—a capability that remains difficult to replicate in robots through behavioral cloning alone, due to the suboptimality and limited diversity of human demonstrations. In this work, we present VT-Refine, a visuo-tactile policy learning framework that combines real-world demonstrations, high-fidelity tactile simulation, and reinforcement learning to tackle precise, contact-rich bimanual assembly. We begin by training a diffusion policy on a small set of demonstrations using synchronized visual and tactile inputs. This policy is then transferred to a simulated digital twin equipped with simulated tactile sensors and further refined via large-scale reinforcement learning to enhance robustness and generalization. To enable accurate sim-to-real transfer, we leverage high-resolution piezoresistive tactile sensors that provide normal force signals and can be realistically modeled in parallel using GPU-accelerated simulation. Experimental results show that VT-Refine improves assembly performance in both simulation and the real world by increasing data diversity and enabling more effective policy fine-tuning.
Our project page is available at \url{https://binghao-huang.github.io/vt_refine/}.

\end{abstract}
\vspace{-10pt}
\keywords{Tactile Simulation, Bimanual Manipulation, RL Fine-Tuning} 


\addtocontents{toc}{\protect\setcounter{tocdepth}{-1}}

\begin{figure*}[!h]
    \centering
    \includegraphics[width=\linewidth]{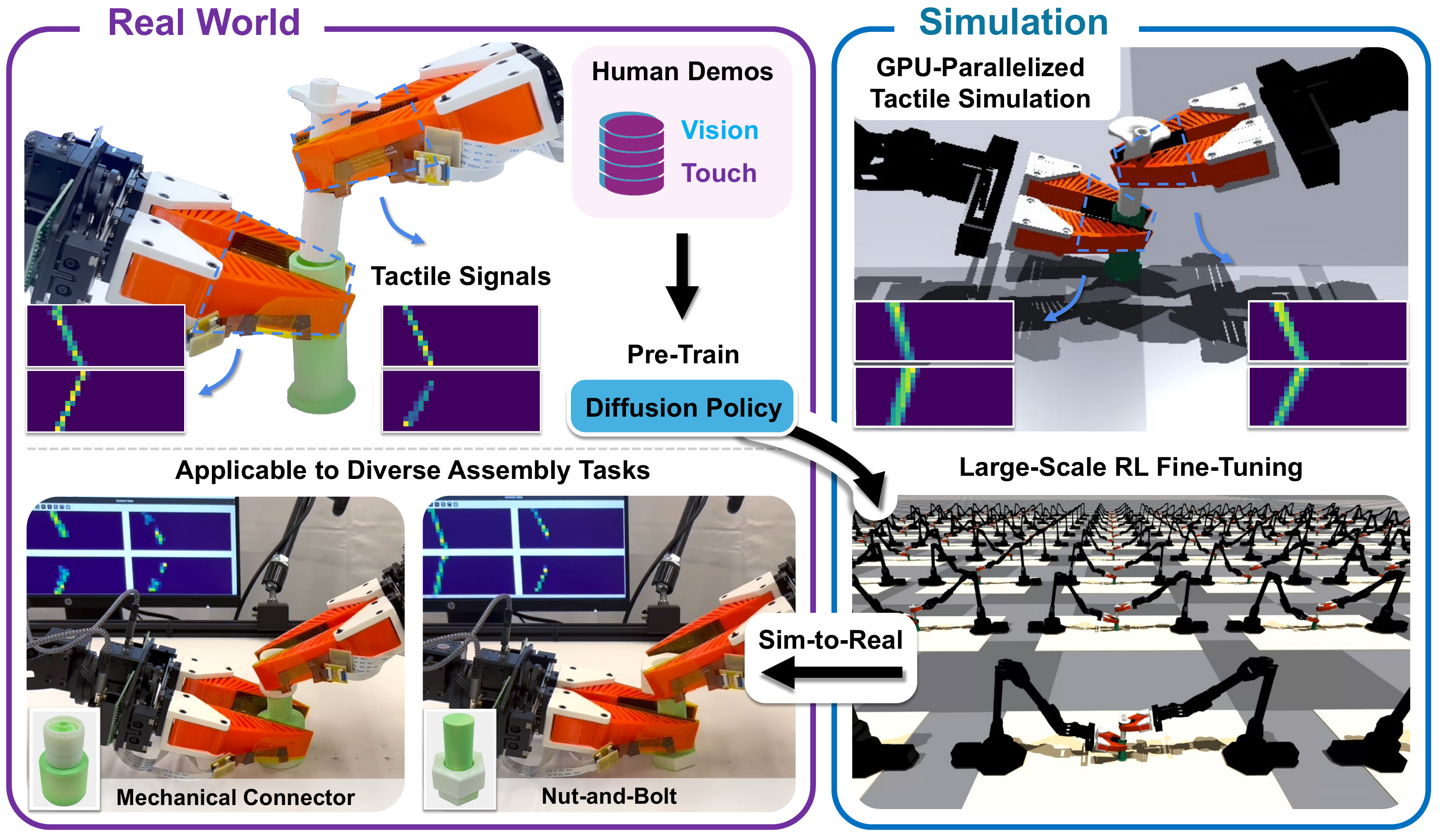}
    \caption{\small{We propose \textbf{VT-Refine}, a novel visuo-tactile policy learning framework for precise, contact-rich bimanual assembly tasks.
\textit{Top Left:} We collect real-world demonstrations and pre-train a diffusion policy using visuo-tactile inputs.
\textit{Right to Bottom Left:} We leverage tactile simulation and large-scale reinforcement learning to fine-tune the policy, and subsequently transfer the fine-tuned policy back to the real world. The resulting policy demonstrates strong performance in both simulated and real environments.
    }}
    
    \vspace{-8pt}
    \label{fig: teaser}
\end{figure*}

\vspace{-5pt}
\section{Introduction}
\label{sec:intro}
\vspace{-5pt}
Solving precise assembly tasks with both hands requires sophisticated orchestration of vision and tactile sensing. Consider a bimanual plug-and-socket assembly: humans first rely on vision to locate the parts and coordinate the grasping and pickup of each part with both hands. Once the parts are held and positioned for insertion, tactile feedback becomes essential. This is because contact cues can be visually occluded during insertion (Fig.~\ref{fig: teaser}), and hence vision alone lacks the precision needed for fine-grained, contact-rich interactions.

Behavioral cloning with diffusion policies~\cite{chi:rss2023,ze:rss2024} has recently shown promise in learning bimanual visuo-tactile policies from a limited number of human teleoperated demonstrations~\cite{huang:corl2024,lin:icra2025}. However, scaling these methods for high-precision assembly tasks in real-world settings poses two major challenges. First, collecting real-world demonstrations is costly, and the demand for data only grows with increasing task precision and contact complexity, making large-scale collection prohibitively expensive. Second, current demonstration interfaces often lack tactile feedback, hindering the capture of how humans use touch for fine manipulation. Consequently, the collected demonstrations typically omit exploratory behaviors---such as iterative adjustments---that are critical for contact-rich tasks, resulting in suboptimal training data. Alternatively, simulation offers a promising path to scale visuo-tactile policy learning, but existing efforts primarily focus on visual modalities or tasks with limited reliance on touch~\cite{villasevil:rss2024,ren:iclr2025,ankile:icra2025}. While some recent work has explored simulation-based data collection for tactile inputs, these efforts are typically restricted to simpler tasks or setups (e.g., unimanual)~\cite{yin:rss2023,qi:corl2023,yuan:icra2024,wang:corl2024,akinola:tro2025,yin:icra2025}, or have not yet addressed large-scale training or robust sim-to-real transfer for tactile-critical bimanual tasks~\cite{lin:ral2023}.

To address these challenges, we introduce a novel real-to-sim-to-real framework designed for precise bimanual assembly. Our approach begins by collecting a small amount of real-world demonstrations (e.g., 30 episodes) to pre-train a bimanual visuo-tactile diffusion policy. The policy is subsequently fine-tuned using reinforcement learning (RL) on a digital twin of the scene within a parallelized simulation environment. Finally, the fine-tuned policy is transferred from simulation back to the real world. Our framework offers three key contributions: (1) We enhance visuo-tactile diffusion policies through RL-based fine-tuning in simulation, enabling policy improvement by exploring state-action regions near those seen in the initial human demonstrations. (2) We develop a GPU-parallelized tactile simulation module within a GPU-based physics simulator to accurately simulate piezoresistive tactile sensors that reliably capture normal force signals. This selection for tactile modality and simulation significantly narrows the sim-to-real gap and overcomes critical challenges in tactile modality transfer. (3) We adopt point-based representations for visual and tactile modalities, facilitating a seamless real-to-sim-to-real transfer. The unified representation preserves the spatial relationships between visual and tactile points, enhancing policy effectiveness. To the best of our knowledge, our work is the first to show successful RL with large-scale simulation and sim-to-real transfer for bimanual visuo-tactile policies.

We comprehensively evaluate our system on five challenging bimanual assembly tasks, demonstrating successful real-world execution and performance gains from simulation-based fine-tuning. A detailed analysis of each training phase shows that high-resolution tactile feedback significantly boosts policy effectiveness during both pre-training and fine-tuning. Additionally, our visuo-tactile point-based representation enables robust bidirectional transfer between real and simulated environments, playing a critical role in the success of our two-stage training framework across tasks and domains.

\vspace{-5pt}
\section{Related Work}
\vspace{-5pt}
\label{sec:related}
\textbf{Tactile Sensors and Simulation.}
Tactile information is critical in human daily life and plays an equally important role in enabling robots to interact with their environments~\cite{Johansson2004RolesOG}. Recognizing its importance, researchers have integrated vision and tactile sensing to enhance robotic manipulation~\cite{yuan:sensors2017,yin:rss2023,yin:icra2025,huang:corl2024,lin:corl2024,suresh2023neural,lin:icra2025,lin:ral2023,lee2019making,dave2024multimodal,10.3389/fnbot.2023.1181383,xue2025reactive,ai2024robopack,bhirangi2024anyskin,guzey2023see}.
Most existing work focuses on optical-based tactile sensors, which can capture normal and shear forces, as well as fine-grained surface textures~\cite{yuan:icra2024,akinola:tro2025,donlon2018gelslim,taylor2022gelslim,ma2019dense,si2024difftactile,softbubble}. However, the  high-resolution images produced by these sensors are difficult to simulate accurately and cause larger sim-real gap. Some approaches~\cite{sunil2023visuotactile,she2021rss,wang2021wedge} attempt to sample marker positions from optical tactile images and infer normal and shear forces from marker deviations, but this indirect method further complicates sim-to-real transfer.
In contrast, we select a tactile sensing modality that emphasizes structural contact patterns with normal force only rather than fine textures. Such signals are not only easier to simulate accurately but also more amenable to transfer between real and simulated environments, enabling scalable visuo-tactile data generation through simulation.

\textbf{Bimanual Visuo-Tactile Manipulation.}
Bimanual robotic manipulation presents significant challenges across a range of applications~\cite{huang2023dynamic,lin:corl2024,wang2023mimicplay,wang2024dexcap,qin2023anyteleop,ankile:icra2025,jiang2024transic}, particularly for assembly tasks. Recently, there has been growing interest in learning-based methods, such as imitation learning~\cite{chi2024universal,wang2024gendp,ankile2024juicer,yu2023mimictouch,zhu2025touch}, which leverage multimodal human demonstrations for fine-grained manipulation. However, achieving higher precision tasks vastly increases the amount of training data required in bimanual settings. To address this, simulation has been used to generate additional data and enhance policy robustness. Villasevil et al.~\cite{villasevil:rss2024} explored the use of reinforcement learning to fine-tune policies initialized by imitation learning.
Nonetheless, most bimanual manipulation frameworks are still restricted to using the visual input alone~\cite{cheng2024tv,lu2024pmp,ding2024bunny}, particularly for real-to-sim-to-real pipelines. This is largely because tactile signals, especially those from optical tactile sensors, are difficult to simulate and transfer~\cite{su2024sim2real}, limiting their potential of being incorporated into simulation-based training.
In contrast, our framework, along with the selection of a transfer-friendly tactile modality, enables effective real-to-sim-to-real learning with both vision and tactile inputs.

\vspace{-5pt}
\section{Visuo-Tactile System and Tactile Simulatiom}
\label{sec:system}
\vspace{-5pt}
\begin{figure*}[!t]
    \centering
    \includegraphics[width=\linewidth]{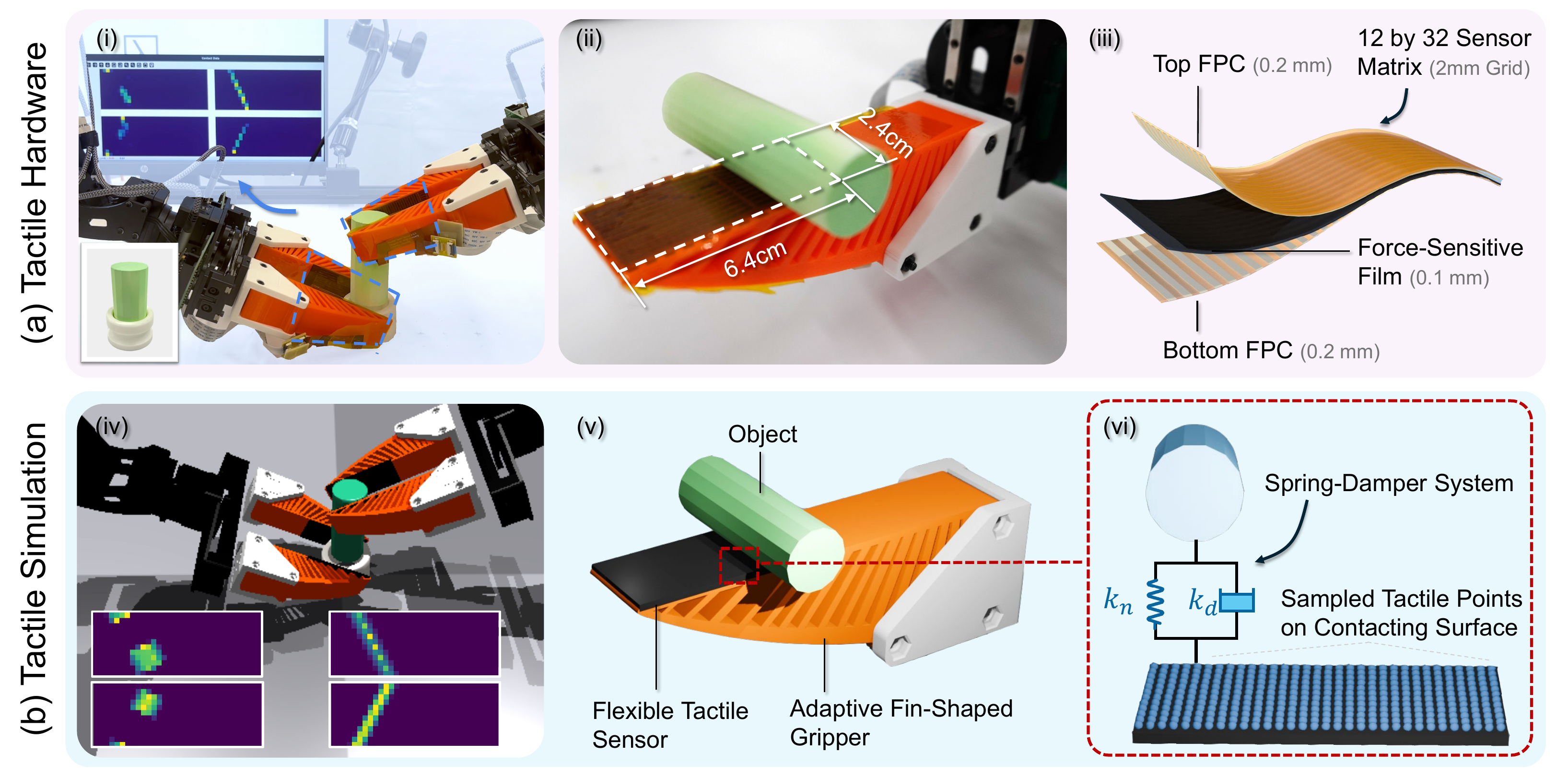}
    \vspace{-15pt}
    \caption{\small{\textbf{Tactile Sensing in Real and Simulation.} (a) Our real-world hardware setup, including the design of the piezoresistive tactile sensor. Four tactile sensor pads (two per hand) are mounted on the soft gripper to capture contact forces. (b) Replication of the tactile sensing process in simulation. A spring-damper model is used to simulate the interaction between the tactile points and objects to generate realistic tactile signals.
    }}
    \vspace{-10pt}
    \label{fig: system}
\end{figure*}

\textbf{Tactile Sensor Hardware.} Our sensor \textbf{FlexiTac} uses resistive sensing matrices, inspired by 3D-ViTac~\cite{huang:corl2024}, to efficiently convert mechanical pressure into electrical signals. The choice of matrix-based flexible sensors is motivated by two key factors:
(1)~\textit{Compatibility}: these sensors are versatile and can be mounted on a wide range of robotic end-effectors, including both rigid and compliant fingers.
(2)~\textit{Sim-to-real transferability}: the sensing modality can be simulated with high fidelity in our environment, enabling consistent behavior across real-to-sim-to-real transfers.

As depicted in Fig.~\ref{fig: system}, each robotic finger is equipped with a tactile sensor pad composed of 12$\times$32 sensing units, with a spatial resolution of 2$mm$ (i.e., 2$mm$ center-to-center distance between adjacent sensors). We use a triple-layer structure similar to~\cite{huang:corl2024}, with a piezoresistive sensing layer sandwiched between two flexible printed circuit (FPC) layers (Fig.~\ref{fig: system} (iii)). Utilizing FPC significantly enhances spatial consistency and increases the resolution of each sensor pad. Additionally, we developed a streamlined fabrication process capable of producing a single sensor in under 5 minutes, enabling cost-effective and scalable deployment. \textbf{We are committed to releasing comprehensive tutorials detailing the hardware design and fabrication process.}

\textbf{Tactile Simulation.} 
To simulate the tactile sensory input, we build on TacSL~\cite{akinola:tro2025}, a GPU-based tactile simulation library integrated with Isaac Gym~\cite{makoviychuk:neuripstdb2021}.
We chose TacSL since the sensory signals acquired from our sensor pads are closely akin to TacSL's simulated tactile normal forces. To model the soft-contact interactions between our deformable tactile sensors (mounted on the soft grippers) and rigid contacting objects, we follow TacSL and employ a penetration-based tactile force model~\cite{xu:corl2022}. 
As shown in Fig.~\ref{fig: system} (vi), the interaction between each \textit{tactile point} (i.e., the 3D position of a sensing unit) and the rigid object is modeled using a Kelvin-Voigt model, consisting of a linear spring and a viscous damper connected in parallel. Each sampled tactile point independently computes the contact normal force $f_n$ as: $f_n = -(k_nd+ k_d\dot{d}) \textbf{n}$,
where $d$ and $\dot{d}$ represent the interpenetration depth and the relative velocity along the contact normal, respectively, and $\textbf{n}$ denotes the outward contact normal vector. At each simulation timestep, the signed distance field (SDF) of the contacting object is queried to compute $d$, and the positions of tactile points are updated in real time via forward kinematics. The known resolution of our tactile sensor allows us to uniformly distribute contact points across the sensor surface. The shape and spatial resolution of the simulated sensor are fully customizable, ensuring consistency with their real-world counterparts. Further implementation details and force computation steps are provided in the Appendix.

\textbf{Real-to-Sim-to-Real for Tactile.}
Alternative tactile sensors, such as vision-based ones like GelSight~\cite{yuan:sensors2017}, rely heavily on internal illumination, making them difficult to simulate accurately and prone to sim-to-real gaps. In contrast, our approach uses normal force signals, which are easier to calibrate in simulation, as demonstrated in our experiments. Another challenge lies in simulating the deformable nature of flexible tactile sensors. While high-fidelity techniques like finite element methods (FEM) can model this softness, they are computationally expensive and impractical for large-scale reinforcement learning. By leveraging TacSL’s GPU-accelerated simulation, we efficiently approximate the softness of flexible tactile sensors, enabling scalable training. As a result, our sensor design improves robustness and supports effective zero-shot sim-to-real transfer.

\vspace{-5pt}
\section{Visuo-Tactile Policy Policy Optimizatiom}
\label{sec:method}
\vspace{-5pt}
\begin{figure*}[!t]
    \centering
    \includegraphics[width=\linewidth]{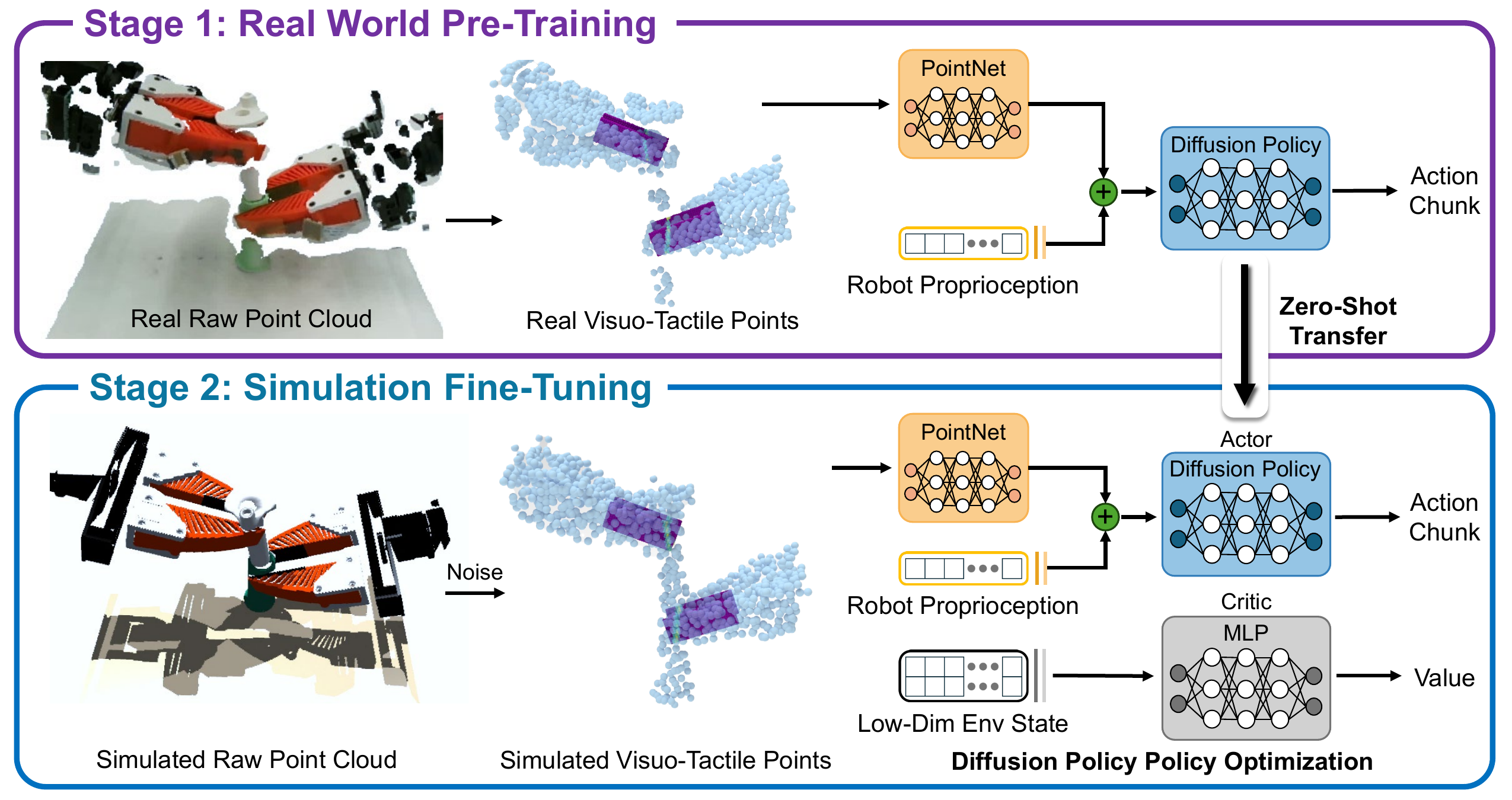}
    \vspace{-17pt}
    \caption{\small{\textbf{Two-Stage Visuo-Tactile Policy Training.}  
\textit{Stage 1:} We collect real-world human demonstrations with visual and tactile modalities and pre-train a diffusion policy.  
\textit{Stage 2:} We simulate the same sensory modalities in simulation and fine-tune the pre-trained diffusion policy using policy-gradient-based RL.
    }}
    \vspace{-13pt}
    \label{fig: method}
\end{figure*}

Our goal is to learn a generalizable and robust control policy, denoted as $\pi: \mathcal{O} \rightarrow \mathcal{A}$ that maps multi-modal observation  $o \in \mathcal{O}$ to robot actions $a \in \mathcal{A}$ with a few real-world demonstrations. As shown in Fig.~\ref{fig: method}, our method consists of two stages: (1)~\textit{real-world pre-training} and (2)~\textit{simulation fine-tuning}. In the first stage, we pre-train a diffusion policy~\cite{chi:rss2023} using behavioral cloning on a small amount of human demonstrations. This pre-trained policy is expected to succeed on the task sporadically for a restricted range of object initial positions in both real-world and simulated environments. In the second stage, we initialize the policy (actor) model from the pre-trained weights and further optimize it with policy-gradient-based RL~\cite{schulman:arxiv2017} in simulation. Finally, this fine-tuned policy is transferred back to the real world for evaluation.

\textbf{Visuo-Tactile Representation.}
The choice of observation $o$ is crucial for bridging the simulation and real world. We adopt a point cloud-based representation for its robust sim-to-real transferability~\cite{qin:corl2022,yuan:icra2024,huang:corl2024}. Our observation contains three modalities: (1)~\textit{visual:} a colorless point cloud captured by an ego-centric camera, denoted as $P^\text{visual}_{t}\in \mathbb{R}^{N_\text{vis}\times 4}$, (2)~\textit{tactile:} a point cloud derived from the tactile sensors representing the 3D positions of the sensing units and their continuous sensory readings, denoted as $P_t^\text{tactile}\in\mathbb{R}^{N_\text{tac}\times 4}$. We set $N_\text{tac}=384\times N_\text{finger}$ for the tactile point cloud since each sensor pad consists of $12\times32=384$ tactile points, and (3)~\textit{proprioception:} joint positions from the two arms and two grippers. 

As shown in Fig.~\ref{fig: method}, we merge the visual and tactile point clouds into a unified visuo-tactile representation: $o = P_t^\text{tactile} \cup P^\text{visual}_{t}$. The tactile sensor's position is computed via forward kinematics and transformed into the camera’s 3D coordinate frame, preserving the spatial relationships between the two modalities. Following~\cite{ze:rss2024}, the merged point cloud is processed by a PointNet encoder~\cite{qi:cvpr2017}, and its output is concatenated with proprioceptive features encoded by a multilayer perceptron (MLP). The resulting feature vector is used as the conditioning input for the denoising diffusion network.

\textbf{Stage 1: Real-World Pre-training.}
We begin by collecting a small real-world demonstration dataset (e.g., 30 episodes in our experiments) to pre-train a diffusion policy. At the beginning of each trial, the assembly parts are randomly placed within a designated region on the table. A human operator then teleoperates both robot arms to pick up the parts and complete the assembly task. During each demonstration, we record visual and tactile inputs, robot joint states, and action commands. To train the diffusion policy, we adopt a denoising diffusion probabilistic model (DDPM) and follow standard practice by predicting an action chunk~\cite{chi:rss2023} (Fig.~\ref{fig: method}, top). Given the limited number of demonstrations, the trained model may not succeed consistently, but is expected to occasionally complete the task, providing reward signals for reinforcement learning to further improve the policy during fine-tuning.

\textbf{Stage 2: Simulation Fine-tuning.}
We fine-tune the pre-trained diffusion policy in an end-to-end manner using Diffusion Policy Policy Optimization (DPPO)~\cite{ren:iclr2025} (Fig.~\ref{fig: method}, bottom). DPPO optimizes a diffusion policy using Proximal Policy Optimization (PPO)~\cite{schulman:arxiv2017}, by formalizing the denoising process as a Markov Decision Process (MDP), which allows the reward signal to propagate effectively through the denoising chain. For scalable training, we assume access to a digital twin of the scene equipped with simulated vision and tactile sensors. The pre-trained diffusion policy initializes the actor network, while the critic network is initialized randomly. We adopt an asymmetric actor-critic strategy~\cite{pinto:rss2018}, where the critic receives a low-dimensional representation of the robot and object state.
\textbf{Reward Function.} In line with the observations in DPPO~\cite{ren:iclr2025}, we find that pre-training on human demonstrations provides a strong prior that guides RL exploration, allowing us to avoid complex reward engineering. We therefore fine-tune using a sparse reward: the agent receives a reward of 1 when the parts are successfully assembled, and 0 otherwise~\cite{heo:rss2023}.

\vspace{-5pt}
\section{Experimental Results}
\label{sec:result}
\vspace{-5pt}
In this section, we address the following three questions through experiments:
(1) How does our fine-tuned policy improve over the baseline diffusion policy?
(2) How effectively does the proposed visuo-tactile representation transfer across domains (real-to-sim-to-real)?
(3) How does policy performance scale with the number of human demonstrations?

\vspace{-5pt}
\subsection{Tactile Simulation Calibration}
\vspace{-5pt}
To align the simulated tactile response with that of the real sensor, we first characterize the sensor’s force-reading curve using a \textit{DMA 850 Dynamic Mechanical Analyzer}.
We then fit a Kelvin–Voigt viscoelastic model by iteratively tuning the elastic modulus $k_n$ (compliance stiffness) and viscosity coefficient $k_d$ (damping) until the simulated curve closely matches the measured response. To validate the calibration, we grasp objects from multiple poses in the real world and record the corresponding tactile signals. We then replay the same trajectories in simulation to collect synthetic tactile data. A histogram comparison of the two datasets shows that the calibrated simulator accurately reproduces the distribution of real tactile signals (Fig.~\ref{fig: calibration}). We provide detailed calibration procedures in the supplementary materials.

\begin{table}[t]
    \tiny
\begin{minipage}{0.43\linewidth}
        \centering
    \includegraphics[width=\linewidth]{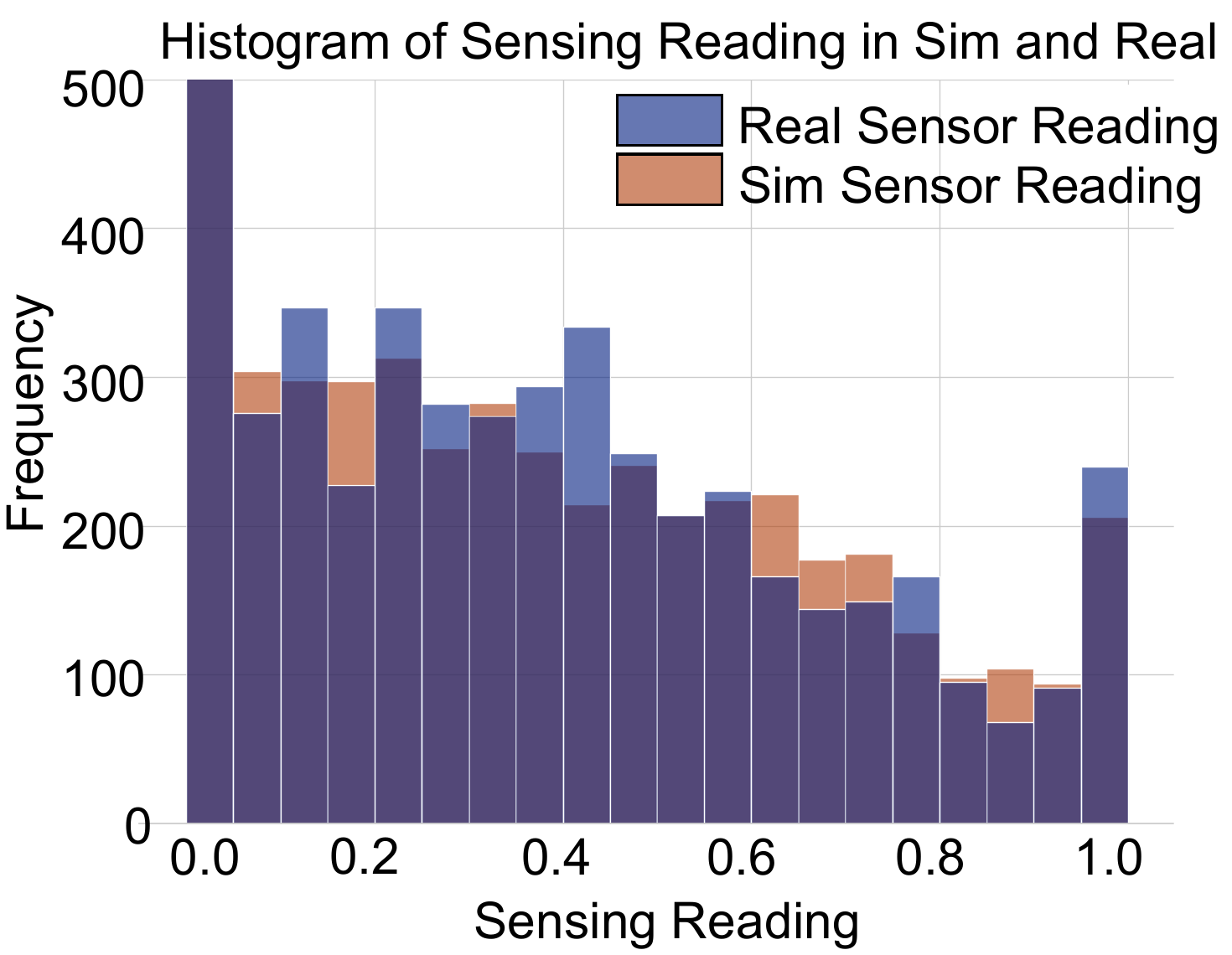}
    \vspace{0.5pt}
    \captionof{figure}{\small{\textbf{Sensor Calibration Results.} The histogram shows consistent sensor reading distributions between the simulation and real world.}}    
    \vspace{-8pt}
    \label{fig: calibration}

\end{minipage}
\hspace{0.01\linewidth}
\begin{minipage}{0.57\linewidth}

        \centering
    \includegraphics[width=\linewidth]{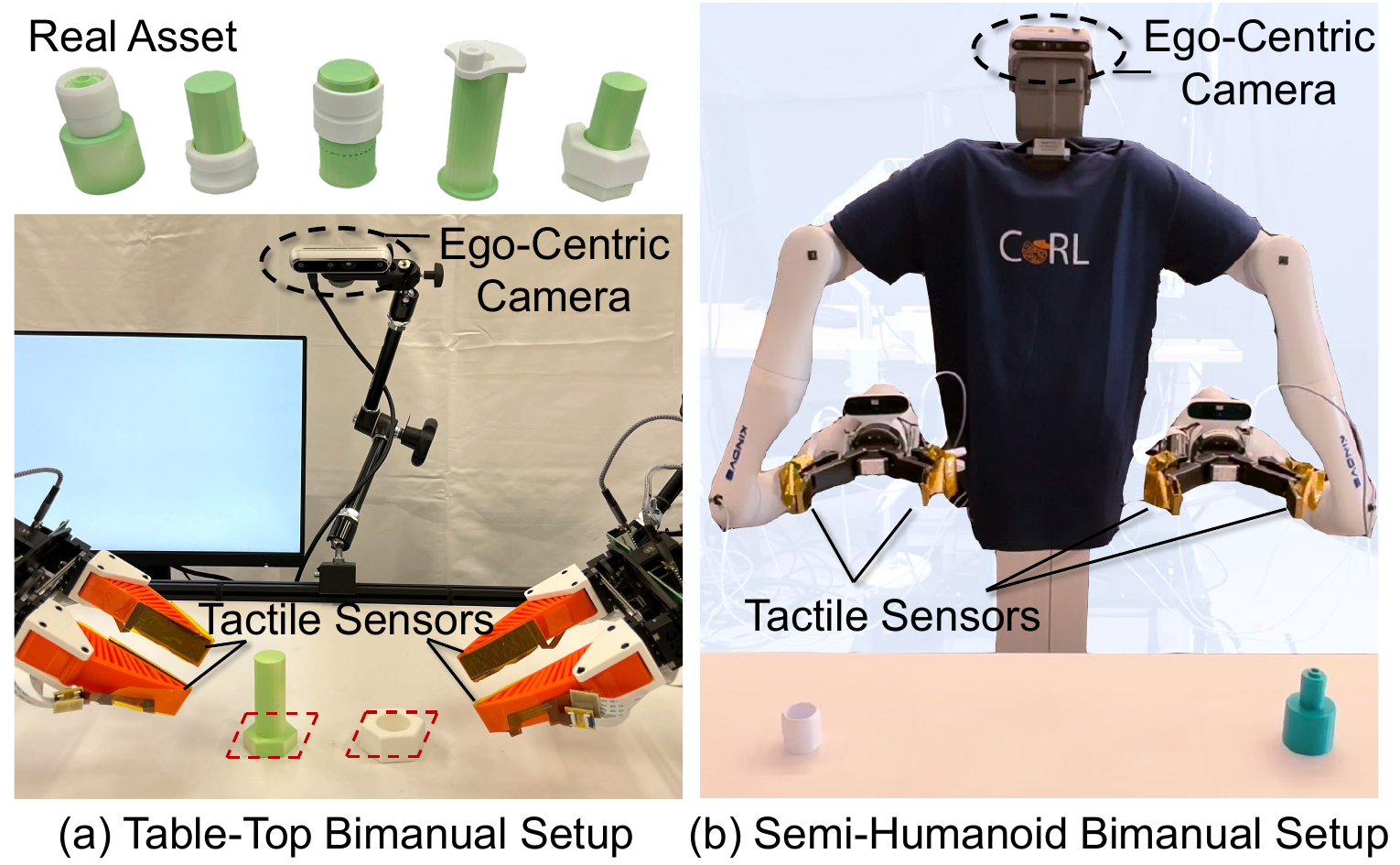}
    \vspace{0.5pt}
    \captionof{figure}{\small{\textbf{Real Robot Setups.} Both the socket and plug are randomly placed in a $3cm \times 3cm$ area. Both robot setups have four tactile sensing pads and an ego-centric camera.}}
    \label{fig: robot}
\end{minipage}
\vspace{-20pt}
\end{table}

\vspace{-5pt}
\subsection{Experiment Setup}
\vspace{-5pt}
We evaluate our multimodal real-to-sim-to-real system on challenging bimanual assembly tasks.
Since our tactile sensors and simulation pipeline can be easily transferred across different robot platforms, we evaluate our method on two setups:
(1) a tabletop bimanual robot arm setup, and
(2) a semi-humanoid robot.
For the \textbf{tabletop bimanual setup} (Fig.\ref{fig: robot} (a)), we adopt the teleoperation setup proposed in ALOHA~2~\cite{aldaco2024aloha}, using two 6-DoF WidowX robotic arms for manipulation, each equipped with a fin-shaped parallel soft gripper. A separate pair of identical arms is used for teleoperation. An Intel RealSense D455 camera is mounted on the table for egocentric visual sensing, and a tactile sensing pad is installed on each of the four soft fingers.
For the \textbf{semi-humanoid setup} (Fig.~\ref{fig: robot} (b)), we use two 7-DoF Kinova Gen3 arms, each paired with a Robotiq 2F-140 gripper. The arms are mounted to a static torso structure. An Intel RealSense D455 camera is mounted at the head for visual sensing, and a tactile sensing pad is attached to each of the four gripper fingers.
For teleoperation, we use the Meta Quest~2, with tracked controller poses mapped to target poses for the robot end effectors. Online trajectory generation is performed using the GPU-accelerated model predictive control framework provided by cuRobo~\cite{sundaralingam:arxiv2023}.

Tasks are selected from the \textit{AutoMate} dataset~\cite{tang:rss2024}. Each task involves a plug-socket pair: the robot must grasp both objects and complete an in-air insertion. Figure~\ref{fig: robot} shows the objects and robot configurations used in our experiments. For each task, we collect 30 demonstrations to pre-train a diffusion policy, which is then used to initialize the policy for fine-tuning (as described in Sec.~\ref{fig: method}). To reflect the variability introduced by bimanual insertion, we randomize the initial pose of each object within a 3$cm$ range during both data collection and fine-tuning. The same visuo-tactile representation and encoder architecture are used throughout pre-training and fine-tuning.

\begin{table}[t]
    \tiny
\begin{minipage}{0.74\linewidth}
    \centering
    \includegraphics[width=\linewidth]{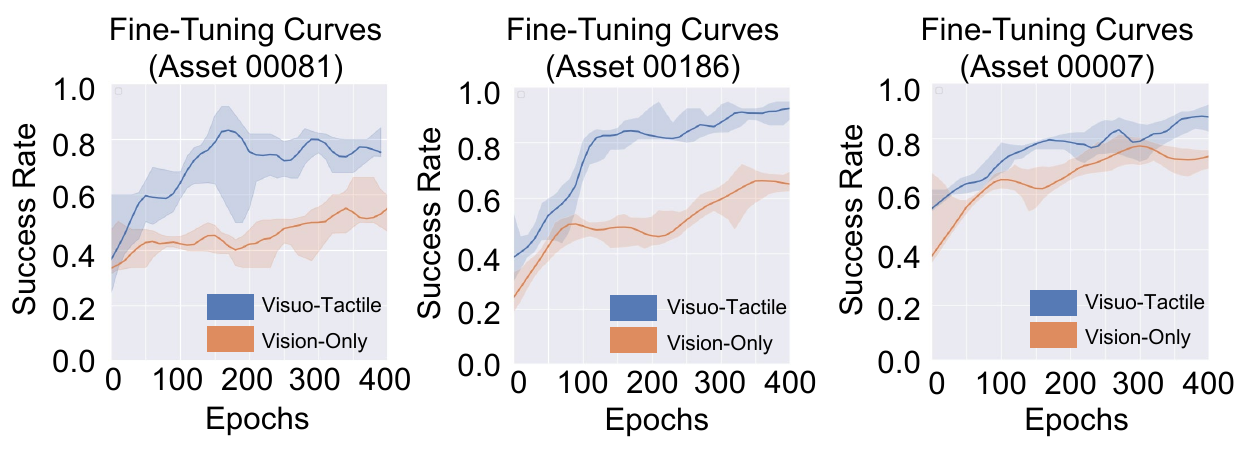}
    \captionof{figure}{\small{\textbf{Simulation Fine-Tuning of Pre-Trained Policies.} We compare the fine-tuning performance of visuo-tactile (blue) and vision-only (orange) policies. The visuo-tactile policy starts with not only a higher pre-trained performance but also continues to improve, achieving higher final performance after fine-tuning.
    }}
    \label{fig: training curve}

\end{minipage}
\hspace{0.01\linewidth}
\begin{minipage}{0.23\linewidth}
    \centering
    \includegraphics[width=\linewidth]{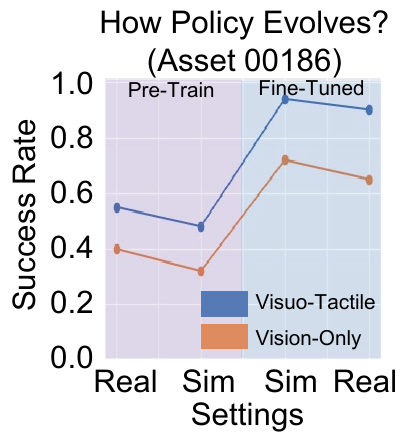}
    \vspace{0.005pt}
    \captionof{figure}{\small{Performance of \textit{Pre-Trained Policy} in sim and real, \textit{Fine-Tuned Policy} in sim and real.
    }}
    \label{fig: performance_curve}
\end{minipage}
\vspace{-20pt}
\end{table}

\begin{figure*}[!ht]
    \centering
    \includegraphics[width=\linewidth]{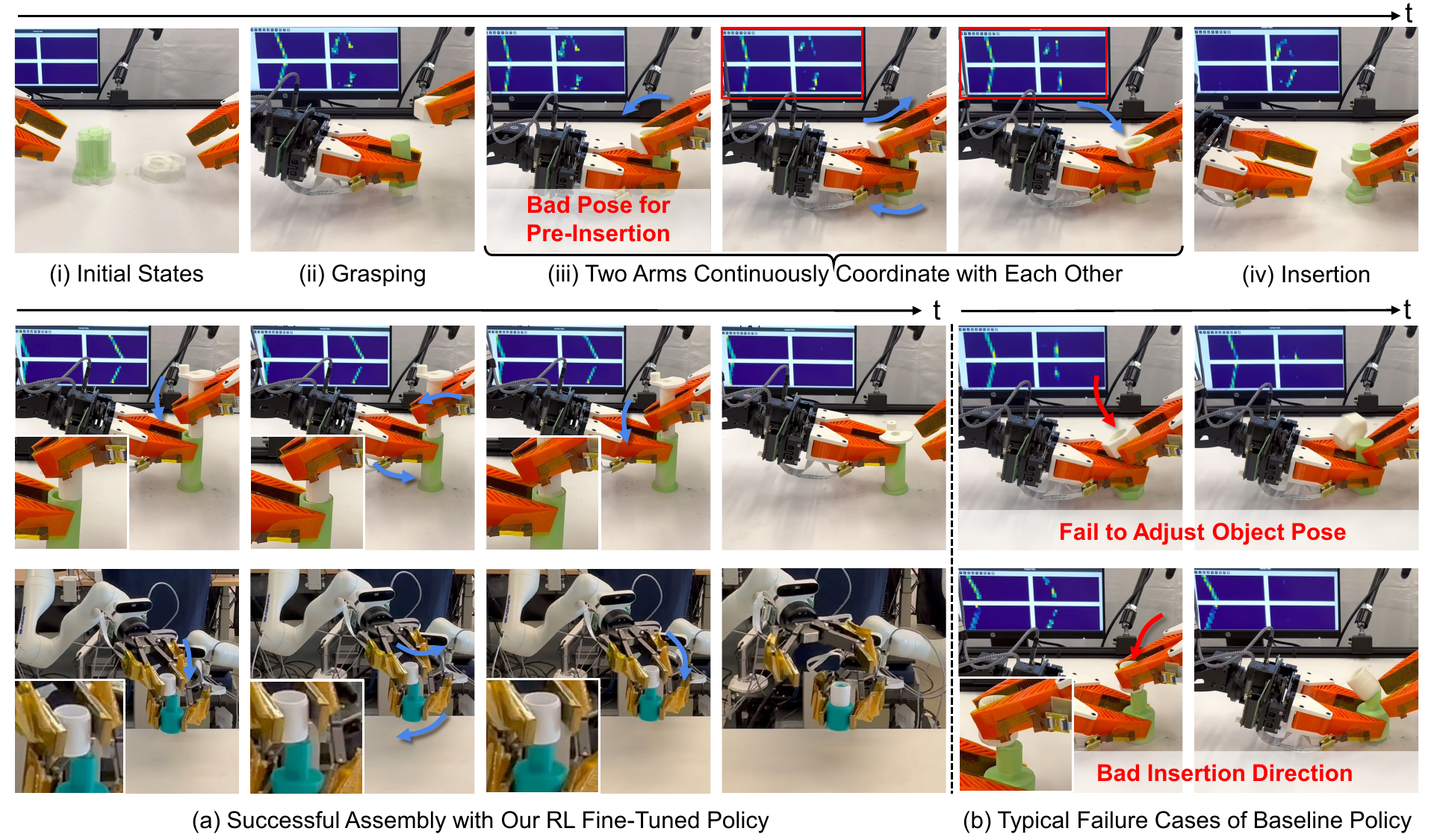}
    \vspace{-15pt}
    \caption{\small{\textbf{Policy Rollout.} We evaluate our fine-tuned visuo-tactile policy on five plug-and-socket pairs with a clearance of roughly $\approx 2mm$. \textit{Part(a)} shows the two arms keep re-orienting or moving the parts until they slide together smoothly, as indicated by the evolving tactile maps on the screen. \textit{Part(b)} the robot either stops with a misaligned pose or pushes at a bad angle, leading to jams and incomplete insertions.
    }}
    \vspace{-15pt}
    \label{fig: exp_vis}
\end{figure*}

\begin{table}[t]
    \centering
    \resizebox{\textwidth}{!}{%
      \begin{tabular}{@{}c@{}}        
        \begin{tabular}{l|ccccc|ccccc}
          \multicolumn{11}{c}{Table-Top Bimanual Setup} \\ \hline
          \multirow{2}{*}{\textbf{Settings}}
            & \multicolumn{5}{c}{\textbf{Visual Policy} (Real)} \vline
            & \multicolumn{5}{c}{\textbf{Visuo-Tactile Policy} (Real)} \\
            & 00081 & 00186 & 00007 & 00446 & 00581
            & 00081 & 00186 & 00007 & 00446 & 00581 \\ \hline
          Pre-Train        & 0.35 & 0.40 & 0.40 & 0.20 & 0.35 & 0.55 & 0.55 & 0.65 & 0.40 & 0.35 \\
          RL Fine-Tuning   & \underline{0.50} & \underline{0.65} & \underline{0.75} & \underline{0.30} & \underline{0.45} & \textbf{0.85} & \textbf{0.90} & \textbf{0.95} & \textbf{0.80} & \textbf{0.75}
        \end{tabular}
        \\[8pt]                         
        \begin{tabular}{l|ccc|ccc}
          \multicolumn{7}{c}{Semi-Humanoid Robot Setup} \\ \hline
          \multirow{2}{*}{\textbf{Settings}}
            & \multicolumn{3}{c}{\textbf{Visual Policy} (Real)} \vline
            & \multicolumn{3}{c}{\textbf{Visuo-Tactile Policy} (Real)} \\
            & 00081 & 00186 & 00007 
            & 00081 & 00186 & 00007 \\ \hline
          Pre-Train        & 0.15 & 0.25 & 0.35  & 0.30 & 0.30 & 0.35  \\
          RL Fine-Tuning   & \underline{0.35} & \underline{0.30} & \underline{0.45}  & \textbf{0.60} & \textbf{0.65} & \textbf{0.65} 
        \end{tabular}
      \end{tabular}%
    }

    \vspace{3pt}
    \caption{\small\textbf{Real-World Experiments.}
      We compare the pre-trained diffusion policy with the policy after RL fine-tuning, as well as vision-only versus visuo-tactile representations. Five object assets are evaluated across two robot setups, with each column corresponding to an \textit{AutoMate}~\cite{tang:rss2024} asset ID.}
    \label{tab: real_exp}
    \vspace{-15pt}
\end{table}

\begin{table}[t]

    \centering
    \resizebox{1\textwidth}{!}
    {
    \begin{tabular}{l|ccccc|ccccc}
        \multicolumn{11}{c}{Table-Top Bimanual Setup}                                       \\ \hline
        
         \multirow{3}{*}{\textbf{Settings}}& \multicolumn{5}{c}{\textbf{Visual Policy} (Sim)}  \vline  & \multicolumn{5}{c}{\textbf{Visuo-Tactile Policy} (Sim)}  \\
          & 00081& 00186  & 00007  & 00446   & 00581  & 00081 & 00186  & 00007  & 00446  & 00581\\
\hline
        Pre-Train  & 0.28 & 0.32 & 0.42 & 0.12& 0.18 & 0.45  & 0.48 &  0.54  & 0.34 &0.31 \\
        
        Fine-Tune w/o Pretrain  & 0.00 &  0.00 & 0.00& 0.00 & 0.00 & 0.00 &  0.00 & 0.00& 0.00 & 0.00  \\
        Fine-Tune w/ Pretrain & \underline{0.57} & \underline{0.72} & \underline{0.84} & \underline{0.36} & \underline{0.52} & \textbf{0.82} &\textbf{0.94}  & \textbf{0.98}  &\textbf{0.76} & \textbf{0.78}\\
    \end{tabular}
    }
    \vspace{3pt}
    \caption{\small{\textbf{Simulation Results.} 
    We compare three variants in simulation: the \textit{Pre-Train Policy} (trained only on real-world demonstrations), \textit{Fine-Tune w/o Pre-Train}, and \textit{Fine-Tune w/ Pre-Train}. The results indicate that both pre-training and fine-tuning contribute significantly to final performance.
    }}
    \label{tab: sim_exp}
    \vspace{-15pt}
\end{table}

\begin{table}[!h]
\vspace{3pt}
    \centering
    \resizebox{0.7\textwidth}{!}
    {
    \begin{tabular}{l|cc|cc}
    \hline
        
         \multirow{3}{*}{Num of Pretrain Data}& \multicolumn{2}{c}{\textbf{Visual Policy} (Sim)}  \vline  & \multicolumn{2}{c}{\textbf{Visuo-Tactile Policy} (Sim)}  \\
          & Pretrain &RL Fine-Tune   & Pretrain & RL Fine-Tune  \\
\hline
        10 demonstrations  & 0.08 &0.21 & 0.02& 0.34 \\

        30 demonstrations  & 0.40 & 0.65 & 0.48 & 0.94 \\

        50 demonstrations  &0.37 &0.67 & 0.57 & 0.92 \\

    \end{tabular}
    }
    \vspace{3pt}
    \caption{\small{\textbf{Different Amount of Pre-Training Data.} We train \textit{Pre-Train Policies} with different amounts of real-world data and transfer them to the simulation for fine-tuning. The results are only compared in simulation.
    }}
    \label{tab: ablation}
    \vspace{-20pt}
\end{table}

\vspace{-5pt}
\subsection{Quantitative Analysis}
\vspace{-5pt}

\textbf{Fine-tuning improves precise manipulation.}
Our RL fine-tuned policy significantly boosts performance on high-precision assembly tasks.
\textit{(i) Fine-tuning introduces necessary exploration.}
Diffusion Policy performs well on lower-precision tasks, but behavior cloning alone lacks the small, repeated adjustments needed for tight-fit insertions. Encoding these subtle exploratory behaviors via demonstrations would require prohibitively large datasets. In contrast, RL fine-tuning introduces such behaviors efficiently by leveraging simulated rollouts.
In real-world experiments (Tab.~\ref{tab: real_exp}), our fine-tuned policy improves success rates by approximately \(20\%\) for the vision-only variant and \(40\%\) for the visuo-tactile variant. Grasping often induces slight object slip, yielding an uncertain pre-insert pose that vision alone seldom detects due to occlusion. Fig.~\ref{fig: exp_vis} (a) shows a representative success trajectory: following an imprecise pre-insertion pose, the two arms engage in rapid cycles of sensing, micro-adjusting, and re-sensing. These back-and-forth ``wiggle-and-dock'' maneuvers—commonly used by humans—emerged organically during RL fine-tuning, despite not being explicitly captured in demonstrations. This is because tactile feedback clearly signals when alignment improves, as indicated through the change of the contact forces.
In contrast, policies without tactile input or sufficient exploration tend to stall or attempt insertion at poor angles, leading to failure or physical damage (Fig.~\ref{fig: exp_vis} (b)).

\textit{(ii) Tactile feedback enhances policy fine-tuning.}
 Visual input alone often fails to capture the fine contact cues needed to align parts. By incorporating tactile data, the visuo-tactile policy gains access to these subtle interactions, enabling it to start from a stronger baseline after pre-training and achieve higher precision after fine-tuning.
As shown in our simulation experiments (Tab.~\ref{tab: sim_exp}), both the vision-only and visuo-tactile policies benefit from fine-tuning. However, the visuo-tactile policy not only begins at a higher performance level but also converges to greater precision. A common failure mode for the vision-only baseline is stalling with the plug hovering just above the socket, unable to close the final 2$mm$ gap. In contrast, the visuo-tactile policy continues adjusting until successful insertion is achieved.

\textbf{Representation transfer across domians (real-to-sim-to-real).}
Transferring a policy between the real robot and simulation inevitably introduces some performance loss due to domain mismatch. These discrepancies arise from differences in point cloud inputs, tactile readings, robot controller settings, and minor joint encoder errors (which affect the placement of tactile points, as they are computed from joint states). As shown in Fig.~\ref{fig: performance_curve}, even with our low-gap tactile modality, we observe a slight performance drop: transferring from real to simulation reduces success rates by approximately $5–10\%$, while sim-to-real transfer causes a smaller—and sometimes negligible—drop. However, since RL fine-tuning in simulation improves success rates by over $30\%$, this transfer loss is acceptable and does not outweigh the overall gain.

\textbf{Ablation study: effect of pre-training data quantity.}
We trained three base policies using 10, 30, and 50 demonstrations, and applied the same RL fine-tuning procedure to each. As shown in Tab.~\ref{tab: ablation}, the policy trained with only 10 demonstrations performed poorly, achieving near-zero success. However, RL fine-tuning still improved its success rate to around $30\%$. The base policies trained on 30 and 50 demonstrations achieved reasonable performance and both fine-tuned to near-perfect success rates. Increasing the dataset from 30 to 50 demonstrations led to minimal improvement in the base policy. In both cases, the policy was already capable of completing the grasp phase; the main bottleneck was the fine, real-time adjustments required during the insertion phase. These micro-motions are difficult to capture with a modest increase in demonstration data, so adding more demonstrations brought limited additional benefit.


\vspace{-5pt}
\section{Conclusion}
\vspace{-5pt}
\label{sec:conclusion}

In this paper, we present a real-to-sim-to-real pipeline with multi-modal perception for precise bimanual manipulation. We introduce a tactile simulation capable of effectively modeling dense tactile sensing grids, achieving strong alignment between simulation and the real world. Finally, we demonstrate the effectiveness of RL finetuning, which substantially improves performance across diverse precise assembly tasks.


\vspace{-5pt}
\section{Limitations}
\vspace{-5pt}

\subsection{Trade-offs with Vision-Based Tactile Sensors}

We compare our \textbf{FlexiTac} sensor with vision-based tactile sensors along three dimensions: 

\textit{(1) Resolution.} Vision-based tactile sensors can provide high-resolution RGB tactile feedback at sub-millimeter ($<$1$mm$) scales, which benefits fine-grained manipulation. However, this high resolution introduces a significant sim-to-real gap. In contrast, our design, while lower in resolution (2$mm$ per unit), reduces simulation complexity and achieves more reliable sim-and-real alignment, while still supporting compact assembly tasks. 

\textit{(2) Shear Force.} A major advantage of vision-based tactile sensors is their ability to capture shear-force information. Although FlexiTac does not provide direct shear-force measurements, policies with temporal history can implicitly infer shear-related effects from contact dynamics. 

\textit{(3) System Design.} Vision-based tactile sensors are typically bulky (at least as large as the camera’s focal length) and thus difficult to integrate into compliant grippers or small size fingertips. Moreover, customizing such sensors~\cite{zhao2025polytouch,ma2024gellink,liu2025vitamin} requires significant engineering effort. In contrast, FlexiTac’s thin, flexible force mat is lightweight, easy to install, and readily customizable, making it well-suited for compliant gripper integration. 

\subsection{Methodological Limitations}

\textit{(1) Real-Sim Alignment.} Like sim-to-real pipelines, our real-to-sim transfer pipeline requires manual calibration efforts. Visual domains, tactile distributions, and low-level control must all be aligned. 

\textit{(2) Scope of the Method Applicability.} The current pipeline is constrained by simulator capabilities: both Isaac Gym and our tactile simulation lack support for deformable objects. Longer-horizon tasks require significantly more training time, and the absence of shear force sensing limits applicability to more complex tasks. In future work, we aim to further enhance real-to-sim fidelity.

\textit{(3) Requirement for CAD models.}
Even though our sparse-reward formulation avoids excessive reward engineering, we still rely on object CAD models to train fine-tuned policies; in this work, we used 3D-printed replicas from an existing dataset. A CAD-free, plug-and-play pipeline would enable the extension of our approach to a much broader range of everyday objects.

\vspace{-5pt}
\section*{Acknowledgement}
\vspace{-5pt}
This work is partially
supported by the Toyota Research Institute (TRI), the Sony
Group Corporation, Google, Dalus AI, Pickle Robot, and an Amazon Research Award. This article solely
reflects the opinions and conclusions of its authors and should
not be interpreted as necessarily representing the official
policies, either expressed or implied, of the sponsors.



\bibliography{main}  
\clearpage

\appendix

\begin{center}
    \LARGE \bf Supplementary Materials
\end{center}

\addtocontents{toc}{\protect\setcounter{tocdepth}{2}}
\tableofcontents

\label{sec: supp}
\section{Details for Tactile Sensor Hardware}

Our \textbf{FlexiTac} sensor builds on prior work~\cite{huang:corl2024,Sundaram2019LearningTS} with several key modifications. 
Instead of manually aligned conductive yarn, we employ \emph{flexible PCBs}, which significantly improve durability and scalability, and we further adapt the design for seamless integration with soft-fin grippers. 
The sensor operates within a force range of approximately $0.2$--$10\,\mathrm{N}$. 
Each sensor pad costs about \$10, while the upgraded reading board costs roughly \$30. 
Compared to the earlier version~\cite{huang:corl2024}, the new reading board now supports up to $32 \times 32$ sensor units at $23\,\mathrm{Hz}$, doubling the resolution from the previous $16 \times 16$ limit. 
The core innovations of this version lie in the \emph{flexible PCB pad design} and the openly released circuit drawings and manufacturing guide, which are available on the \href{https://flexitac.github.io/}{FlexiTac website}.

\section{Details for Tactile Simulation Implementation}

\begin{figure*}[!ht]
    \centering
    \includegraphics[width=\linewidth]{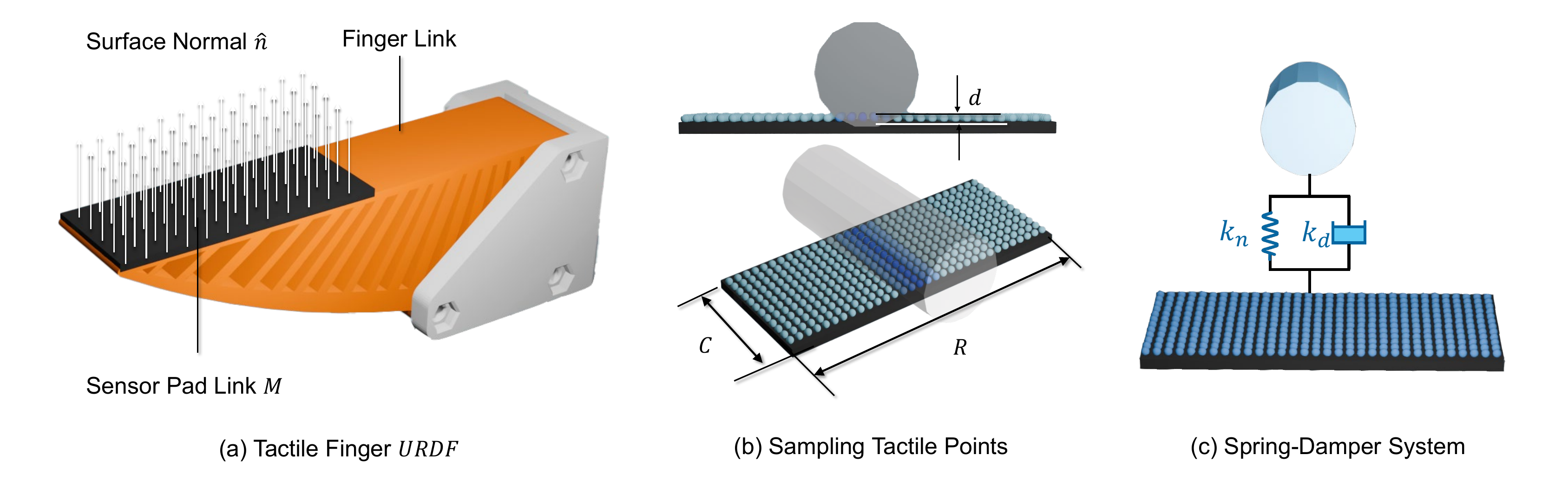}
    \vspace{-15pt}
        \caption{\small\textbf{Tactile‑Simulation Pipeline.} 
        \textit{(a)} Finger model in the simulator with a dedicated elastomer sensing pad. 
        \textit{(b)} Uniform grid of taxels sampled on the pad surface and their contact interaction with an external object. 
        \textit{(c)} Penalty‑based spring–damper model applied at every taxel to convert penetration into normal‑force and depth signals.}
    \vspace{-15pt}
    \label{fig: system_supp}
\end{figure*}

\subsection{Sampling Tactile Points}

As shown in Fig.~\ref{fig: system_supp}, each robot finger carries a sensor pad defined as a separate mesh in the URDF. Let $M$ denote the triangular mesh of this pad and $\hat{\mathbf n}$ its outward surface normal (toward the ``finger tip'' link).

\begin{itemize}
    \item $M$ be the triangle mesh of the sensor pad link,
    \item $\hat{\mathbf{n}}$ the local surface normal that points towards the
          internal compliant layer (the ``tip'' link in the URDF).
\end{itemize}

Given a desired resolution $R\times C$ (the real robot uses $12\times32$), we generate taxel positions in three consecutive steps. 

\textit{(i) Detect the flat face:} the shortest bounding‑box axis of $M$ corresponds to pad thickness; the two remaining axes span the contact surface.

\textit{(ii) Create a planar lattice:} A rectilinear grid is laid over this face, leaving a 1 mm margin to avoid edge artifacts; the grid spacing is equal to $d_{\text{taxel}}$.  

\textit{(iii) Ray‑cast for ground‑truth locations:}  
rays are shot from every lattice node along $-\hat{\mathbf n}$ until they
intersect $M$.  The resulting 3‑D points populate \texttt{tactile\_pos\_local}$\in\mathbb{R}^{N\times3}$ ($N=R\times C$) and are all assigned the fixed orientation $q_{\text{taxel}}=\mathrm{Euler}(0,0,-\pi)$ so that the $+y$ axis always points outward in world space.

This fully automatic procedure yields a dense, uniform taxel layout and requires no manual annotation.

\subsection{Tactile Signal Computation}

At every physics step we convert geometric contacts into a dense
\textbf{two‑channel} image that the policy ingests directly:

\begin{center}
\begin{tabular}{ccl}
\toprule
\textbf{Chan.} & \textbf{Symbol} & \textbf{Quantity (tactile frame)} \\
\midrule
0 & $d$      & Penetration depth (m) \\
1 & $f_{n}$  & Normal force (N)      \\
\bottomrule
\end{tabular}
\end{center}

\textit{(i) Taxel pose in world coordinates.}\;
For every taxel~$i$
\[
    \mathbf x_i^{w}= \mathbf R_e\,\mathbf x_i^{l}+\mathbf p_e,
    \qquad
    \dot{\mathbf x}_i^{w}= \boldsymbol\omega_e\!\times\!
                            (\mathbf R_e\mathbf x_i^{l})+\mathbf v_e ,
\]
where $(\mathbf R_e,\mathbf p_e,\boldsymbol\omega_e,\mathbf v_e)$ are the pose
and twist of the sensor pad link returned by PhysX.

\textit{(ii) SDF query.}\;
A single GPU kernel returns the signed distance $d_i$, outward normal
$\hat{\mathbf n}_i$, and normal relative velocity
$\dot d_i=\hat{\mathbf n}_i\!\cdot\!\dot{\mathbf x}_i^{w}$ for every taxel.

\textit{(iii) Penalty contact model.}\;
The normal contact force is
\[
    f_{n,i}=-(k_n\, d_i + k_d\, \dot d_i),
\]
with constants $k_n = 1.0$ and $k_d = 3\times10^{-3}$.
Shear forces are not used in our cases.

\textit{(iv) Packing and normalisation.}\;
Negative depths ($d_i<0$, meaning no contact) are clamped to zero.
The pair $(d_i,f_{n,i})$ is linearly rescaled to $[0,1]$ and reshaped into an
$R\times C\times4$ tensor that is streamed directly to the policy network.

The loop is fully vectorised over all environments (\texttt{N\_envs}) and fully GPU-parallelized.

\section{Real-to-Sim-to-Real}

\subsection{Tactile Calibration}

To match simulated tactile readings to the real hardware we adjust only the normal stiffness \(k_n\) and the damping term \(k_d\) of the penalty model. Calibration follows three steps: \textit{(i)} a single taxel on the finger pad is chosen in both domains; \textit{(ii)} a sequence of forces is applied to the real pad, yielding a ground‑truth force–response curve; and \textit{(iii)} the same normal loads are replayed in simulation while \(k_n\) and \(k_d\) are iteratively tuned until the mean‑squared error between the two curves is minimised. Then apply this parameter to all sensor units.

Real pads exhibit a small, load‑independent noise floor.  We reproduce this behaviour by a two‑stage normalisation
\[
  s^{\text{norm}} =
  \begin{cases}
     s / s_{\max}^{\text{fixed}}, & s < \tau,\\[4pt]
     s / s_{\max}^{\text{curr}},  & s \ge \tau ,
  \end{cases}
\]
where \(s\) is the raw taxel reading, \(\tau\) is a noise threshold,
\(s_{\max}^{\text{fixed}}\) is a constant from the sensor data‑sheet, and
\(s_{\max}^{\text{curr}}\) is the frame‑wise maximum over the pad.
The identical rule is applied to simulated readings so that both domains share
the same dynamic range.

\begin{figure*}[!ht]
        \centering
        \includegraphics[width=0.5\linewidth]{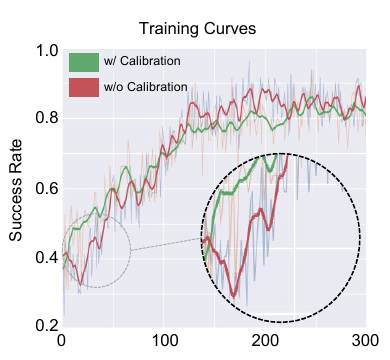}
        \caption{\small\textbf{Effect of Tactile Calibration on RL Fine‑Tuning.} Training curves with and without sensor‑simulation calibration. Without calibration the success rate initially falls as the policy re‑adapts to the shifted tactile distribution, whereas the calibrated variant improves monotonically from the first iteration.}
        \vspace{-15pt}
        \label{fig: compare_curve_supp}
\end{figure*}

As illustrated in Fig. 4 of main text, the histograms of normalised signals overlap almost perfectly after calibration.
The benefit is further demonstrated in Fig.~\ref{fig: compare_curve_supp}, which shows fine‑tuning performance with and without calibration using the \emph{same} Pre-Trained policy (trained by real data). The calibrated run improves steady up, whereas the uncalibrated run initially degrades as the policy adapts to the shifted observation distribution. Although both eventually succeed in simulation, the uncalibrated policy exhibits a larger sim‑to‑real gap when deployed on the robot, underscoring the importance of distribution matching.

\subsection{Vision for Sim–Real Alignment}

To minimise the domain gap between simulated and real observations we normalise the depth stream from a single egocentric RealSense camera through three successive operations.

\textit{(i) Point‑cloud generation:} raw depth values are back‑projected to camera space and transformed into the world frame whose origin coincides with the table centre.

\textit{(ii) Workspace cropping:} the cloud is clipped to a manually specified axis‑aligned bounding box that covers the robot’s reachable workspace and excludes background clutter.

\textit{(iii) Uniform down‑sampling:}
to accelerate both data loading and on‑line RL roll‑outs we subsample the cloud to a fixed budget of points using uniform lin‑space indexing. Although farthest‑point sampling (FPS) is common in diffusion‑policy pipelines, uniform sampling is \(\sim\!10\times\) faster in our setting and was found to have no measurable impact on task performance.

\textit{(iv) Noise injection (simulation only):} To emulate RealSense depth jitter, each simulated point \(\mathbf x_w\) is perturbed by a multiplicative factor,
\(
  \mathbf x_w \leftarrow \mathbf x_w
  \bigl(1 + \mathcal{N}(0,\,0.01\,\sigma)\bigr),
\)
where \(\sigma\) is a user‑controlled noise level. We use noise level 3 for our pipeline. This step is omitted for real‑camera data.

\section{DPPO Implementations and Parameters}

\subsection{Pre-Train Implementations and Parameters}

\textbf{Diffusion model.}
We employ the epsilon‑prediction variant of \emph{Diffusion Policy}~\cite{chi2024diffusionpolicy} with \(T\!=\!100\) denoising steps and a rollout horizon of
\(H\!=\!16\) actions. Conditioning information is injected during the first \(C\!=\!2\) steps for proprioception and during the first \(C_{\text{img}}\!=\!2\) steps for point clouds, following the two‑stream scheme in~\cite{chi2024diffusionpolicy}.

\textbf{Backbone.}  

\textit{(i) Input structure.}
At each conditioning step~\(t\) the policy receives merged visuo-tactile point cloud:

\begin{itemize}[nosep]
\item \emph{Visual cloud} \(\mathcal P^{\,v}_t\in\mathbb{R}^{3\times N_v}\), whose fourth channel is initialised to zeros.
\item \emph{Tactile cloud} \(\mathcal P^{\,\tau}_t\in\mathbb{R}^{4\times N_\tau}\)  containing XYZ positions of taxels in the world frame and their normalised pressure readings as a fourth channel.
\end{itemize}

To allow the network to distinguish modalities we append a one‑hot flag, yielding a 5‑channel tensor  
\[
\mathbf P_t = 
\bigl[
  (\mathcal P^{\,v}_t;\,\mathbf 0),~
  (\mathcal P^{\,\tau}_t;\,\mathbf 1)
\bigr]
\in\mathbb{R}^{5\times(N_v+N_\tau)} ,
\]
which is transposed to the \((N,5)\) format expected by PointNet.

\textit{(ii) Per‑step backbone.}
Each \(\mathbf P_t\) is processed by
\emph{PointNetEncoderXYZ\_Tactile}, an MLP with hidden sizes
\(\{64,128,256,512\}\).
Optionally, layer normalisation is inserted after every linear
layer; we use the variant with \texttt{LayerNorm} and
final projection to a 64‑d feature.
A global \(\max\nolimits_{n}\) pooling over points produces the
per‑step vector \(\mathbf f_t\in\mathbb{R}^{64}\).

\textit{(iii) State feature.}
If proprioceptive state is available the \(16\)-d joint states at each step is mapped through a two‑layer MLP (\(\{64,64\}\)) and concatenated with \(\mathbf f\). The joint proprioception can be obtained both in sim and real.

\subsection{Fine-Tune Implementations and Parameters}
\paragraph{Actor network.}
The backbone is identical to pre‑training (PointNet 64 → U‑Net \(\{512,1024,2048\}\), kernel 5, group norm 8), but only the last \(T_{\text{ft}}\) diffusion steps are optimised. A small K‑L penalty (\(\lambda=2{\times}10^{-4}\)) on the predicted noise keeps the decoder close to the pre‑trained manifold.

\paragraph{Critic.} A state‑value network receives the proprioceptive history (concatenated over the \(C{=}2\) conditioning steps, dimensionality \(30\times2\)) and passes it through an MLP (512–512–512, Mish activations, residual connections).

\paragraph{PPO hyper‑parameters.} We collect one segment of
\(n_{\text{steps}}
  =\lfloor240 / 8\rfloor
  =30\) decisions
from every environment before an update. Discount and GAE factors are
\(\gamma=0.999\) and \(\lambda=0.95\). Ten PPO epochs are run per batch with a target K‑L of 1.0. Learning rates are \(10^{-5}\) (actor) and \(10^{-3}\) (critic) with cosine decay to \(10^{-6}\) and \(10^{-3}\), respectively. In the paper DPPO~\cite{dppo2024}, they use learning rate \(5\times10^{-5}\), but here we use \(10^{-5}\) to ensure a more stable training as the task is even more fine-grained. Table~\ref{tab:diffusion_hp} shows the important hyperparamter we truned for our fine-grained manipulation task.

\begin{table}[h]
\centering
\small
\begin{tabular}{lll}
\toprule
Symbol & Value & Config Key \\
\midrule
$\gamma_{\text{denoise}}$      & 0.99       & \texttt{gamma\_denoising}                \\
$\lambda_{\text{KL}}$          & $2\times10^{-4}$ & \texttt{clip\_ploss\_coef} \\
--- base value                 & $2\times10^{-4}$ & \texttt{clip\_ploss\_coef\_base} \\
--- annealing rate             & $3$         & \texttt{clip\_ploss\_coef\_rate}        \\
$\sigma_{\text{rand}}$         & 3.0         & \texttt{randn\_clip\_value}             \\
$\hat{\sigma}_{\min}$          & 0.01        & \texttt{min\_sampling\_denoising\_std}   \\
$\tilde{\sigma}_{\min}$        & 0.10        & \texttt{min\_logprob\_denoising\_std}    \\
\bottomrule
\end{tabular}
\caption{DPPO–specific hyper‑parameters used during fine‑tuning.}
\label{tab:diffusion_hp}
\end{table}

\end{document}